\documentclass[letterpaper, 10 pt, conference]{ieeeconf}  

\IEEEoverridecommandlockouts                              

\usepackage{graphicx} 
\usepackage{epsfig} 
\usepackage{times} 
\usepackage{amsmath} 
\usepackage{amssymb}  
\usepackage{dsfont}
\usepackage{algorithm}
\usepackage{algorithmic}
\usepackage{array,multirow}
\usepackage{booktabs}
\usepackage{pifont}
\usepackage{xcolor}
\usepackage{url}

\title{\LARGE \bf
Radar Enlighten the Dark: Enhancing Low-Visibility Perception for Automated Vehicles with Camera-Radar Fusion
}

\author{Can Cui, Yunsheng Ma, Juanwu Lu and Ziran Wang
\thanks{Can Cui, Yunsheng Ma, Juanwu Lu and Ziran Wang are with Lyles School of Civil Engineering,
        Purdue University, West Lafayette, IN 47906, USA. 
        Email: \{\tt\small cancui, yunsheng, juanwu, ziran@purdue.edu\}}%
}

\begin{document}

\maketitle%
\thispagestyle{empty}
\pagestyle{empty}

\begin{abstract}
Sensor fusion is a crucial augmentation technique for improving the accuracy and reliability of perception systems for automated vehicles under diverse driving conditions. However, adverse weather and low-light conditions remain challenging, where sensor performance degrades significantly, exposing vehicle safety to potential risks. Advanced sensors such as LiDARs can help mitigate the issue but with extremely high marginal costs. In this paper, we propose a novel transformer-based  3D object detection model ``REDFormer'' to tackle low visibility conditions, exploiting the power of a more practical and cost-effective solution by leveraging bird's-eye-view camera-radar fusion. Using the nuScenes dataset with multi-radar point clouds, weather information, and time-of-day data, our model outperforms state-of-the-art (SOTA) models on classification and detection accuracy. Finally, we provide extensive ablation studies of each model component on their contributions to address the above-mentioned challenges. 
Particularly, it is shown in the experiments that our model achieves a significant performance improvement over the baseline model in low-visibility scenarios, specifically exhibiting a 31.31\% increase in rainy scenes and a 46.99\% enhancement in nighttime scenes.
The source code of this study is publicly available\footnote{\url{https://github.com/PurdueDigitalTwin/REDFormer}}.


\end{abstract}


\section{INTRODUCTION}

Sensor fusion, also known as sensor integration, refers to the process of combining data from various sensors installed in an automated vehicle to obtain a comprehensive and accurate understanding of the surrounding environment~\cite{campbell_sensor_2018}, it is also a crucial component in digital twin technology. \cite{9724183}. This approach allows the vehicle to obtain a more comprehensive view of the environment, critical for making informed decisions and navigating safely. The sensors commonly used in automated vehicle sensor integration include LiDAR, radar, camera, and global navigation satellite system (GNSS). Integrating information from multiple sensors allows automated vehicles to operate more effectively and safely in various driving conditions. 

However, one major challenge in sensor fusion applications is the underperformance of sensors in low visibility environments. Common causes for visibility reduction include adverse weathers, such as rain, snow, and low-light conditions (like driving at night). As a result, automated vehicles that rely on these inaccurate detection results pose substantial risks, which might expedite serious traffic accidents or other devastating consequences. Therefore, there is a need to develop a sensor fusion system that is reliable in low visibility and can work effectively in diverse practical conditions.

A number of existing research prefers fusing LiDAR and camera sensors due to their complementary nature. However, such a solution comes with the potential limitations \cite{liu_robust_2022}:
\begin{itemize}
\item \textbf{Costliness:} LiDAR can be very expensive compared to other sensors on automated vehicles.
\item \textbf{Environmental Sensitivity:} Environmental factors such as rain, snow, or fog can reduce LiDAR's accuracy. 
\item \textbf{High Computations:} LiDAR generates large amounts of data, which can be computationally intensive to process and analyze.
\end{itemize}
On the contrary, radar sensors can effectively detect objects under a broader range of environmental conditions and are generally less expensive than LiDARs. Therefore, it is more reasonable to investigate sensor fusion of radar and camera data to achieve the 3D object detection task in a practical context. Compared to relying solely on radar sensors, camera-radar fusion offers numerous benefits, including improved object detection by combining high-resolution images from cameras with the ability to be unaffected by environmental conditions of radar to detect objects even in low-light or adverse weather conditions. Particularly, advanced machine learning algorithms such as Transformers empower the perception capability of automated vehicles by its computational efficiency and scalability \cite{dosovitskiy_image_2021}, and allowing us to train multi tasks in one joint model \cite{can_structured_2021}. Such systems are more robust and provide a complete awareness of the environment, contributing to avoiding potential hazards and accidents. Moreover, camera-radar fusion systems are more accessible and cost-effective for all vehicles, promoting the overall penetration rate.

This paper proposes a 3D object detection model applying sensor fusion on multi-camera images and multi-radar point clouds in a bird's-eye-view (BEV) perspective. Inspired by achievements in natural language processing, we learn the positional representation of multi-radar point clouds using an embedding mechanism and gated linear filters. We achieve spatial-temporal fusion using the attention mechanism with image embedding, radar point cloud embedding, and BEV positional queries. Additionally, we design a multi-task objective to integrate visibility conditions into the end-to-end training procedure, aiming to enable our model to understand different low-visibility conditions and their corresponding weather and time-of-day (TOD) properties comprehensively.

The main contributions of this paper are summarized as follows:
\begin{itemize}
    \item A novel radar embedding backbone is proposed for camera-radar fusion that significantly improves 3D object detection accuracy compared to using images solely.
    \item A multi-task learning (MTL) paradigm is developed to incorporate weather and TOD information as additional contextual clues during training.
    \item Extensive experiments on a 3D object detection benchmark are conducted. The results demonstrate that our model outperforms the state-of-the-art approaches, especially in adverse weather conditions and low-light environments.%
\end{itemize}


\begin{figure*}[t!]
    \centering
    \includegraphics[width=\textwidth]{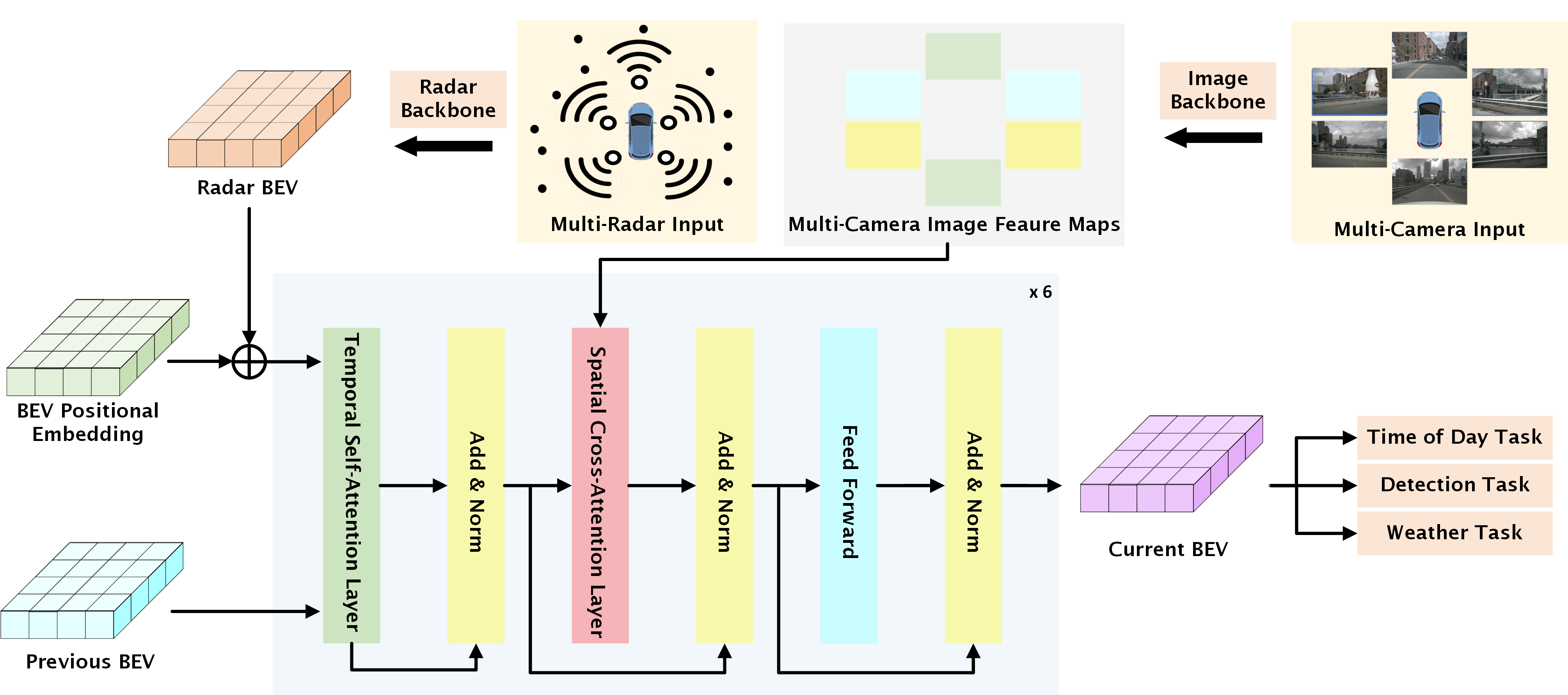}
    \caption{Illustration of the proposed REDFormer. First, the radar backbone generates radar BEV embedding from the multi-radar point cloud. The radar BEV combines with  BEV positional embeddings, and the combination and the learnable BEV features from the previous layer is the input for a temporal self-attention layer. Then, the image backbone extracts multi-view image features. BEV queries from the upstream selectively search within the regions of interest associated with image feature maps in the consecutive spatial cross-attention layer. Finally, a downstream model uses the BEV features from the attention layer to derive predictions that simultaneously minimize multi-task prediction loss.}

    \label{fig:overall_s}
\end{figure*}

\section{Related Works}


\subsection{Object Detection}

Problem of object detection is a computer vision task that aims to identify and localize objects within an image. During the past decade, the developments of machine learning technologies have led to emerging interests in research on 2D object detection, with a multitude of works in this field~\cite{ren_faster_2016,girshick_fast_2015}. Nevertheless, 2D object detection has limitations in representing the depth of view of the context. As a result, more recent research focuses on designing models for 3D object detection. FCOS3D \cite{wang_fcos3d_2021} enhanced a 2D detector FCOS \cite{tian_fcos_2019} and directly generated 3D bounding boxes. Inspired by DETR~\cite{carion_end--end_2020}, DETR3D~\cite{wang_detr3d_2021} extracts 2D features from multiple camera images and associates them with 3D positions using sparse 3D object queries and camera transformation matrices. Using multi-camera image inputs, M$^2$M \cite{xie2022m2bev} achieves 3D object detection and map segmentation in the BEV space. Other methods~\cite{chitta_neat_2021} directly use multi-layer perception to learn how to project images input from the camera to the BEV plane. BEVformer~\cite{li_bevformer_2022} uses a deformable transformer to compute spatial and temporal features in BEV grid regions of interest across camera views and previous BEV information. Besides, existing methods have also investigated how to incorporate radar information to enhance image-based object detection. For example, CRF-Net~\cite{nobis_deep_2020} converts unstructured radar pins into a pseudo-image format and uses a downstream model to process radar embeddings alongside the camera image.


However, many of these studies need to address the performance degradation under low-visibility conditions, which has been attracting attentions recently. Mirza et al. analyzes the performance degradation of different architectures under adverse weather conditions\cite{mirza_robustness_of_object_detectors}. Tomy et al. proposes a robust sensor fusion model that combines event-based and frame-based cameras for robust object detection in adverse conditions, utilizing a voxel grid representation for event input and employs a two-parallel feature extractor network for both frames and events \cite{tomy_fusing_event-based}. Bijelic et al. proposes a deep fusion architecture that enables robust fusion in foggy and snowy conditions \cite{bijelic_seeing_2020}. Sakaridis et al. presents a benchmark dataset comprising authentic images captured in diverse adverse weather conditions \cite{DBLP:journals/corr/abs-1708-07819}. Meanwhile, Kenk provides a benchmark radar dataset for evaluating object-detecting model performance in bad weather \cite{mourad_dawn_vehicle_detection}. Therefore, our work takes one step further to investigate how to exploit multi-radar point clouds to enhance object detection algorithms under low-visibility conditions.

\subsection{Camera-Radar Sensor Fusion}

A previous study by Waldschmidt et al. shows that radar is not affected by adverse lighting and severe weather conditions, enabling direct measurement of distance, radial velocity, and, with the aid of a suitable antenna system, determination of the angle of distant objects \cite{9318758}. Hence, the concept of camera-radar sensor fusion aims to exploit the merits of both types of sensors and provide a more comprehensive understanding of the surroundings. Kadow et al. \cite{kadow_radar-vision_2007} used radar data to limit the search scope for video input and return the distance features and then utilized a simple neural network for vehicle identification. Bertozzi et al. used radar data to refine detecting bounders and apply camera data to detect and classify road obstacles \cite{bertozzi_obstacle_2008}. Other works used similar methods to achieve camera-radar sensor fusion \cite{kocic_sensors_2018, han_frontal_2016}. With the rising deep learning techniques, more efficient sensor fusion models have emerged in recent years. Nobis et al. presented a network model whose layers incorporated both image data and projected sparse radar data and enabled the model to learn features from both sensors \cite{nobis_deep_2020}. \cite{jha_object_2019} and \cite{kim_sensor_2017} involved independent object detection by each sensor (camera and radar), followed by the integration of their results to arrive at a final decision. Lekic et al. projected the radar data onto the 2D images generated by the camera and then used Conditional Multi-Generator Generative Adversarial Networks (CMGANs) to implement object detection \cite{article}. Bijelic et al. developed a deep fusion model using camera and radar inputs and validated it on a dataset with adverse weather conditions \cite{bijelic_seeing_2020}. A middle-fusion method, CenterFusion, proposed in~\cite{nabati_centerfusion_2021} utilized a center point detection network to detect objects based on their center points in the image. Distinguished from the abovementioned studies, our work draws inspiration from recent achievements in natural language processing and uses embedding along with attention mechanisms~\cite{vaswani_attention_2017} to achieve camera-radar sensor fusion.

\section{Methodology}

This section presents our camera-radar bird's-eye-view fusion transformer (REDFormer) model for the 3D object detection in detail. The main design idea is to fuse multi-radar point cloud and multi-camera image features in a bird's-eye-view (BEV) plane with addressing the performance drop under low-visibility conditions. We will start with our problem statement and introduce the key components in our model, including learnable BEV queries, the radar backbone (RB) module, and the multi-task learning (MTL) module that address the objectives in the problem statement.

\subsection{Problem Statement and Model Architecture}

Our model aims to achieve 3D object detection using camera-radar sensor fusion. Suppose we have $N_r$ radar sensors and $N_c$ cameras. At each timestep $t$, we observe of a collection of multi-camera images $\mathcal{F}_t=\left\{f_i^{(t)}\in\mathbb{R}^{3\times H\times W}\mid i=1,\ldots, N_c\right\}$, where $H$ and $W$ are the height and width of each image. Besides, we also observe multi-radar point clouds $\mathcal{P}_t=\left\{p_{ij}^{(t)}\in\mathbb{R}^{D_r}\mid i=1,\ldots, N_r\right\}$, where $j$ is the number of radar points for radar sensor $i$, and $D_r$ the attribute dimension for each radar point. Suppose we have $N_k$ context objects. The bounding box of a context object $k$ is defined by a vector of $D_b$ parameters $b^{(t)}_k\in\mathbb{R}^{D_b}$, and the type of the object is $c^{(t)}_k\in\mathbb{R}$. The objective of our problem is to search for the optimal model inside the model space $\mathcal{M}=\left\{m\mid m:\mathbb{R}^{D_r}\times \mathbb{R}^{3\times H\times W}\rightarrow\mathbb{R}^{D_b}\times\mathbb{R}\right\}$ that minimizes the prediction and classification errors%

\begin{equation}
    \begin{aligned}
        & m^*=\underset{m\in\mathcal{M}}{\text{argmin}}\frac{1}{N_k}\sum\limits_{k=1}^{N_k}\text{err}(b_k^{(t)},\hat{b}_k^{(t)}) + \text{err}(c_k^{(t)}, \hat{c}_k^{(t)}), \\
        \text{where } & (\hat{b}_k^{(t)}, \hat{c}_k^{(t)})=m(\mathcal{F}_t, \mathcal{P}_t\mid\theta).
    \end{aligned}
\end{equation}

\begin{figure*}[t!]
    \centering
    \includegraphics[scale=0.54]{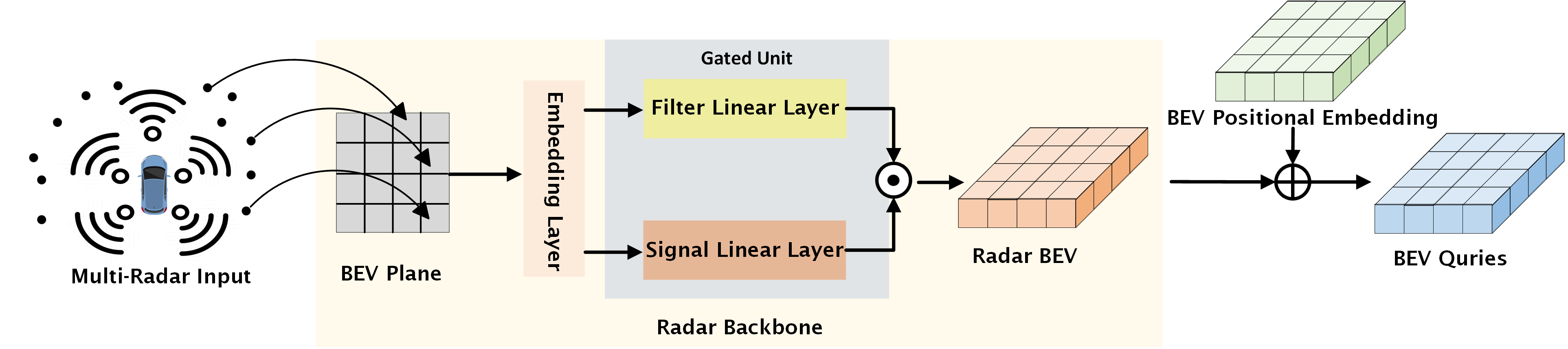}
    \caption{Illustration of the radar backbone in the REDFormer.
The backbone projects multi-radar point clouds onto the BEV plane and then aggregates local point clouds regarding each region of interest (RoI). An embedding layer learns the representation of the saliency signal within each RoI. A gated unit filters the signal and generates the Radar BEV embeddings.}
    \label{fig:radar_backbone}
\end{figure*}

Given the problem statement above, there are three essential problems :
\begin{itemize}
    \item \textbf{Sensor Fusion} How to resolve the differences in feature spaces between multi-radar point clouds and multi-camera images?
    \item \textbf{Temporal Dependencies} How to learn the temporal dependencies between consecutive observations?
    \item \textbf{Vulnerability to Low-visibility Conditions} How to incorporate low-visibility information into the end-to-end learning pipeline?
\end{itemize}

As illustrated in Fig. 1, we approach sensor fusion and temporal dependencies by proposing a novel embedding-based RB and a temporal self-attention layer to unify multi-radar point clouds and multi-camera images in a BEV perspective. The RB handles extracting the positional saliency signal of each region of interest, while the attention mechanism learns temporal correlations between current radar signals and previously observed BEV features. 
 
Nevertheless, the original objective function~\cite{li_bevformer_2022} failed to consider vulnerability to low-visibility conditions. To compensate, we train our model by optimizing a multi-task objective. Specifically, instead of only minimizing the prediction and classification error, our model minimizes the binary classification error of current weather conditions $\hat{w}^{(t)}$ and the time-of-day label (i.e., night or daytime) $T^{(t)}$ simultaneously. The overall architecture of the proposed model is inspired by the existing state-of-the-art model BEVFormer~\cite{li_bevformer_2022}. However, we distinguish ourselves with the three innovative components explicitly designed to address the three critical problems.

\begin{algorithm}[ht!]
    \caption{Train REDFormer}
    \begin{algorithmic}[1]
        \renewcommand{\algorithmicrequire}{\textbf{Input:}}
        \REQUIRE Multi-radar Point $\mathcal{P}_t$, Multi-camera Images $\mathcal{F}_t$ \\ Learning Rate $\eta$, Data $\left\{\{b_k^{(t)}\}, \{c_k^{(t)}\}, w^{(t)}, T^{(t)}\right\}$.
        \STATE Randomly initialize the model $m$ with parameters $\theta$.
        \WHILE {\text{not converged}}
            \STATE $\left(\left\{\hat{b}_k^{(t)}\right\},\left\{\hat{c_k}^{(t)}\right\}, \hat{w}^{(t)}, \hat{T}^{(t)}\right)\leftarrow m(\mathcal{P}_t, \mathcal{F}_t;\theta)$.
            \STATE Compute the detection loss $\mathcal{L}_\text{det}(\hat{b}_k^{(t)}, b_k^{(t)}, \hat{c}_k^{(t)}, c_k^{(t)})$.
            \STATE Compute the weather predict loss $\mathcal{L}_\text{rain}(\hat{w}^{(t)}, w^{(t)})$.
            \STATE Compute the time-of-day predict loss $\mathcal{L}_\text{tod}(\hat{T}^{(t)}, T^{(t)})$.
            \STATE $\mathcal{L}_\text{joint}\leftarrow \mathcal{L}_\text{det}+\mathcal{L}_\text{rain}+\mathcal{L}_\text{tod}$.
            \STATE $\theta \leftarrow \theta + \eta\nabla_\theta\mathcal{L}_\text{joint}$.
        \ENDWHILE
        \RETURN Model $m(\mathcal{P}_t, \mathcal{F}_t\mid\theta)$ 
    \end{algorithmic}
\end{algorithm}

\subsection{Radar Backbone}

\newcommand{\PP}{\mathcal{P}}


It is critical to resolving the difference in feature spaces between multi-radar point clouds and multi-camera image inputs for sensor fusion. To that end, we design the RB module, which projects and helps unify the features under the BEV scope.


Initially, since each radar sensor observes points from its perspective, we need to unify the overall point clouds in a shared local coordinate system centered at the current position of the ego vehicle. For each radar $i$, we project its observed point cloud using an affine transformation matrix ${p^{\prime}}_{ij}^{(t)}=\mathcal{T}_i\cdot p_{ij}^{(t)}$. Then, we aggregate these projected points in a BEV plane by region of interest (RoI) and create a saliency matrix. Suppose we denote the BEV regional saliency matrix by $S\in\mathbb{R}^{X\times Y}$, where $X$ and $Y$ are the numbers of RoI on each row and column, respectively. The value of each cell $s_{ij}\in\mathbb{R}$ is the number of projected multi-radar points presented within the corresponding region of interest.


Drawing our concepts from natural language processing, we propose a novel approach incorporating an embedding layer in conjunction with a gated unit layer for processing the structured radar point data. The embedding layer $E(s_{ij}):\mathbb{R}\rightarrow\mathbb{R}^C $ treats each cell salience $s_{ij}$ as a token and creates a look-up table mapping from the cell salience to a $C$-dimensional learnable vector. We use a gated unit layer alongside the embedding layer to reduce the noise in raw radar saliency and squash the value space. The gated unit consists of a sigmoid-linear function and a tanh-linear function. If to denote the feature in a output Radar BEV cell by $e_{ij}$, we can express the overall operation as follows:
\begin{equation}
    e_{ij} = \sigma(W_1\cdot E(s_{ij}) + b_1) \odot \tanh(W_2\cdot E(s_{ij}) + b_2),
\end{equation}
where $(W_1,W_2)$ and $(b_1, b_2)$ are the weights and biases for sigmoid-linear and tanh-linear functions, and $\odot$ denotes the element-wise product of two matrices.

\subsection{Temporal Self-Attention and BEV Queries}


Object detection models can take advantage of sequential input consisting of previous observations in practical cases. To address the temporal correlations, we implement the temporal self-attention layer with the help of BEV queries. As shown in Fig. \ref{fig:radar_backbone}, the BEV queries matrix $Q\in \mathbb{R}^{X\times{Y}\times{C}}$  derives by adding the radar BEV from RB with a learnable BEV positional embedding. We then fed the BEV queries into the temporal self-attention layer to learn correlations with the BEV representations from preceding timesteps.

Recall that the upstream RB module ensures that the center of the BEV features corresponds to the position of the ego vehicle. Using the intrinsic matrices of cameras, we can determine the corresponding points on the multi-view images for each query $Q_p \in \mathbb{R}^{{C}}$. As a result, we build the connection between points on the multi-view images and points on the BEV queries. Such a connection is vital for fusing radar features with camera features in the consecutive spatial cross-attention layer~\cite{li_bevformer_2022}.

\subsection{Multi-Task Learning}

Multi-task learning (MTL) is an inductive transfer mechanism that aims to enhance generalization performance by leveraging shared information across multiple related tasks \cite{caruana_multitask_1997}. In contrast to traditional machine learning methods, which often focus on learning a single task, MTL capitalizes on the potential richness of information embedded within training signals of related tasks originating from the same domain.

In our proposed REDFormer model, we employ an MTL strategy to bolster its adaptability to diverse environmental circumstances, such as different weather and night scenarios. To achieve this, we apply two additional recognition heads on the unified BEV representation, allowing the model to simultaneously learn and account for weather and TOD conditions alongside the primary 3D object detection task. These recognition heads are designed as separate linear output branches, each responsible for predicting a specific task-related output. The inclusion of these additional learning goals allows the model to capture nuanced relationships and shared features across the tasks. This fosters better generalization and enhanced performance under diverse real-world conditions, resulting in a more robust and versatile 3D object detection system.

Following~\cite{carion_end--end_2020,wang_detr3d_2021}, we employ the set-to-set loss function to quantify the disparity between the predicted and ground truth values for the 3D object detection task. For environmental variables, we utilize the standard cross-entropy loss function to perform binary classification, identifying whether it's raining or not, and whether it's nighttime or not. The joint loss function is defined as $\mathcal{L}_\text{joint}=\mathcal{L}_\text{det}+\mathcal{L}_\text{rain}+\mathcal{L}_\text{tod}$.

\begin{table*}[t!]
    \centering
    \caption{Comparison of the proposed REDFormer with various state-of-the-art methods. The top-performing method for each setting is highlighted in \textbf{bold}. $\downarrow$: Lower values are better. $\uparrow$: Higher values are better. $^*$: Results obtained from the original papers.}
    \begin{tabular}{@{}l|c|cccccccc@{}}
        \toprule
        \multicolumn{1}{c}{Modality} \vrule & Method & \#param. & NDS $(\uparrow)$ & mAP $(\uparrow)$& mATE $(\downarrow)$& mASE $(\downarrow)$& mAOE $(\downarrow)$& mAVE $(\downarrow)$& mAAE $(\downarrow)$\\
        \midrule
        \multirow{5}{*}{Camera only} 
        &FCOS3D$^*$\cite{wang_fcos3d_2021-1} & $\ge$52.5M & 0.415 & 0.343 & 0.725& \textbf{0.263} & 0.422 & 1.292& 0.153\\
        &DETR3D$^*$\cite{wang_detr3d_2021} & 51.3M & 0.425& 0.346 & 0.773 & 0.268 & \textbf{0.383} & 0.842& 0.216 \\
        &BEVFusion$^*$\cite{liu_bevfusion_2022} & - &0.412 & 0.356 & - & - & - & - & -\\
        &BEVDet$^*$\cite{huang_bevdet_2022} & 53.7M& 0.392 & 0.312 & 0.691 & 0.272 & 0.523 & 0.909 & 0.247\\
        &BEVFormer$^*$\cite{li_bevformer_2022} & 56.8M &0.479& 0.370& 0.725 & 0.272 & 0.391 & 0.802 & 0.200\\
        \midrule
        \multirow{2}{*}{Camera and Radar} 
        &CenterFusion$^*$\cite{nabati_centerfusion_2021} & - & 0.453& 0.332 & \textbf{0.649}& 0.263 & 0.535 & 0.540 & \textbf{0.142} \\
        &\textbf{REDFormer (ours)} & 56.8M & \textbf{0.486}  & \textbf{0.385} & 0.726 & 0.282 & 0.407 & \textbf{0.427} & 0.218\\
        \bottomrule
    \end{tabular}
\label{tab:sota}
\end{table*}

\begin{table}[!ht]
\centering
\caption{Comparison of the proposed REDFormer with the best state-of-the-art method in low-visibility subsets. $^*$: Results obtained from the original papers}
\begin{tabular}{@{}l|cccc@{}}
    \toprule
    \multicolumn{1}{c}{Subset} \vrule & Method & NDS $(\uparrow)$ & mAP$(\uparrow)$ \\
    \midrule
    \multirow{3}{*}{Rainy Scenes} 
    & BEVDet$^*$~\cite{huang_bevdet_2022} & - & 0.3370 \\
    & BEVFormer~\cite{li_bevformer_2022}& 0.3877 & 0.3524\\
    & REDFormer (ours)& \textbf{0.5091} & \textbf{0.4036}\\
    \midrule
    \multirow{3}{*}{Night Scenes} 
    & BEVDet$^*$~\cite{huang_bevdet_2022}& - & 0.1350 \\
    & BEVFormer\cite{li_bevformer_2022}& 0.1913 & 0.1819\\
    & REDFormer (ours) & \textbf{0.2812} & \textbf{0.2028} \\
    \bottomrule
\end{tabular}
\label{tab:rainnight}
\end{table}

\section{Experiments}

\subsection{Dataset}

Our model is implemented and evaluated on the nuScenes dataset, a large-scale public dataset designed by Motional for autonomous driving research \cite{caesar_nuscenes_2020}. The nuScenes dataset consists of approximately 1,000 scenes, each with a duration of approximately 20 seconds. For every sample, six high-definition cameras are positioned to capture images at a resolution of 1600 $\times$ 900 pixels covering 360 degree field of view. And five radar sensors provide 360-degree coverage around the vehicle, operating at a frequency of 77GHZ with a range up to 250 meters.

The nuScenes dataset uses a comprehensive set of evaluation metrics to assess the performance of models and algorithms for object detection task. NuScenes dataset uses mean average precision (mAP) as its primary evaluation metric, which compute the 2D center distance on the ground plane for object matching rather than 3D intersection over union affinities. The nuScenes dataset also includes a set of true positive metrics, such as average translation error (ATE), average scale error (ASE), average orientation error (AOE), average velocity error (AVE), and average attribute error (AAE), which are used to evaluate the accuracy of object detection in terms of translation, scale, orientation, velocity, and attribute errors. In addition, the nuScenes dataset introduced a nuScenes detection score (NDS) to consolidate all the above evaluation metrics. The NDS is calculated as follow:
\newcommand{\nds}{\text{NDS}}
\begin{equation}
    \nds = 0.5 \times \text{mAP} + \sum{0.1 \times \max((1-\text{mTP}),0)}
\end{equation}


\subsection{Baseline}

We select several state-of-the-art 3D object detection models as baselines to compare the performance of our approach. The baseline models include DETR3D~\cite{wang_detr3d_2021}, FCOS3D~\cite{wang_fcos3d_2021-1}, BEVFusion~\cite{liu_bevfusion_2022}, BEVFormer (small varient) \cite{li_bevformer_2022} and BEVDet (tiny varient) \cite{huang_bevdet_2022}, which rely solely on cameras. Additionally, we compare our approach to CenterFusion\cite{nabati_centerfusion_2021} and CRF-Net \cite{nobis_deep_2020}, which incorporate both radar and camera data.

\subsection{Main Results}
We conduct extensive evaluations of our proposed REDFormer model on the nuScenes dataset object detection task, and compare its performance with several state-of-the-art object detection approaches that use either camera, radar, or camera-radar fusion models. Our results, summarized in Tab. \ref{tab:sota}, demonstrate that our model achieves a significant improvement over the baselines. Specifically, we obtained a 6.7\% improvement in NDS and a 16.1\% improvement in mAP compared to CenterFusion (camera-radar Fusion) \cite{nabati_centerfusion_2021}, an 1.5\% improvement in NDS and a 4.1\% improvement in mAP compared to BEVFormer (Camera only) \cite{li_bevformer_2022}. These findings indicate the effectiveness of our novel RB and MTL modules, and their abilities to capture the nuances of multi-sensor data and enhance object detection performance in challenging scenarios.

To assess the robustness of our model under adverse weather conditions, we conduct experiments in both low-light nighttime and rainy scenarios. Specifically, we evaluate our model's performance in these adverse conditions to verify its ability to maintain high accuracy and reliability even in low visibility environments. To do this, we create two sub-datasets from the nuScenes dataset, one containing only nighttime scenarios and the other containing only rainy scenes. We then perform experiments on these sub-datasets to evaluate our model's performance in these challenging conditions. We use BEVFormer \cite{li_bevformer_2022} and BEVDet \cite{huang_bevdet_2022} as our baseline model. The object detection performance in low-visibility scenarios is generally not as good as in normal conditions, as indicated in Tab. \ref{tab:rainnight}. However, our REDFormer demonstrates significant improvements in both rainy and nighttime conditions when compared to BEVFormer \cite{li_bevformer_2022}. Specifically, we observe a 31.31\% improvement in NDS on rainy scenes and a 46.99\% improvement in NDS on night scenes highlighting the effectiveness of our approach in adverse weather conditions. Our model also showcases significant improvements than BEVFormer \cite{li_bevformer_2022} in mAP across challenging conditions. Particularly, we achieve a notable 14.5\% improvement in mAP for rainy scenes and an impressive 11.5\% improvement for night scenes. Furthermore, compared to the baseline model BEVDet \cite{huang_bevdet_2022}, our model exhibits a substantial 19.8\% improvement in mAP for rainy scenes and a remarkable 50\% improvement for night scenes. Hence, the notable enhancement in low-visibility scenarios demonstrates the radar's ability to provide precise object localization. This signifies the effectiveness of our RB in guiding the model towards object localization especially in low-visibility conditions, while the MTL heads effectively incorporate light conditions into the predictions. In rainy or nighttime conditions, the model appropriately assigns higher importance to radar inputs compared to sunny or other high-visibility conditions.

\subsection{Ablation Study}
\paragraph{Module analysis} We conduct ablation experiments on the nuScenes validation set to analyze the impact of different modules, specifically the MTL and the RB. Each component is removed individually while the other is kept fixed. Our ablation study includes both the full nuScenes dataset and low-visibility subsets to comprehensively evaluate the effectiveness of our approach. Tab. \ref{tab:ab_m} demonstrates that both the RB and MLT modules significantly improve model performance. Additionally, when compared to the baseline model, Tab. \ref{tab:ab_m_subset} highlights the improvements made by each module in night and rainy scenes. Notably, the RB module improves NDS by 31.21\% and 43.33\% in rainy and night subsets, respectively, while the MLT module improves NDS by 30.51\% and 43.44\% in rainy and night subsets, respectively. Individually, both of these components contribute to enhancing the object detection performance of the model, and their combined usage yields the highest performance, indicating the synergistic benefits of integrating both components in the model. These findings highlight the effectiveness of our proposed approach and underscore the value of the consequential modules in enhancing the performance of object detection systems in challenging scenarios.

\paragraph{Influence of the capacity limit of the embedding dictionary}
We discuss the outcomes of selecting the capacity limit $K$ for the embedding dictionary based on the experiment presented in Tab. \ref{tab:abk}. We restrict our focus to $K\ge10$, as this value represents the maximum number of radar points within a single grid throughout the entire nuScenes dataset. The optimal performance in terms of mAP is achieved when $K=10$, while the NDS values show near identical best performances for $K=10$ and $K=20$. These results suggest that $K$ should be selected as close as possible to the radar point capacity per grid, contingent on the specific radar setup of the vehicle.


\begin{table}[ht!]
\newcommand{\cmark}{\ding{51}}%
\newcommand{\xmark}{\ding{55}}%
\centering
\caption{The performance of REDFormer is compared with and without the inclusion of radar Backbone (RB) and multi-task learning (MTL) in the whole nuScenes dataset. }
\begin{tabular*}{\linewidth}{@{\extracolsep{\fill}}cc|cc@{\extracolsep{\fill}}}
    \toprule
    \multicolumn{2}{c}{Module} \vrule & \multirow{2}{*}{NDS $(\uparrow)$}   & \multirow{2}{*}{mAP $(\uparrow)$}\\
    \cmidrule{1-2}
    With RB & With MTL&\\
    \midrule
    \xmark & \xmark & 0.4787 & 0.3700\\
    \xmark & \cmark & 0.4851 & 0.3838\\
    \cmark & \xmark & 0.4833& 0.3816 \\
    \cmark & \cmark & \textbf{0.4863} & \textbf{0.3853}\\
    \bottomrule
\end{tabular*}
\label{tab:ab_m}
\end{table}

\begin{table}[ht!]
\newcommand{\cmark}{\ding{51}}%
\newcommand{\xmark}{\ding{55}}%
\centering
\caption{ The performance of REDFormer is compared with and without the inclusion of radar Backbone (RB) and multi-task learning (MTL) in rainy and night-time subset.}
\begin{tabular*}{\linewidth}{@{\extracolsep{\fill}}l|cc|cc@{\extracolsep{\fill}}}
    \toprule
    \multicolumn{1}{c}{\multirow{2}{*}{Subset}} \vrule &\multicolumn{2}{c}{Module} \vrule & \multirow{2}{*}{NDS $(\uparrow)$}   & \multirow{2}{*}{mAP $(\uparrow)$}\\
    \cmidrule{2-3}
    &With RB & With MTL&\\
    \midrule
    \multirow{4}{*}{Rainy Scenes} 
    &\xmark & \xmark & 0.3877 & 0.3524\\
    &\xmark & \cmark & 0.5060 & 0.3959\\
    &\cmark & \xmark & 0.5087 & 0.3966\\
    &\cmark & \cmark & \textbf{0.5091} & \textbf{0.4036}\\
    \midrule
    \multirow{4}{*}{Night Scenes} 
    &\xmark & \xmark & 0.1913& 0.1819 \\
    &\xmark & \cmark & 0.2742& \textbf{0.2079} \\
    &\cmark & \xmark & 0.2744& 0.2067 \\
    &\cmark & \cmark & \textbf{0.2812} & 0.2028\\
    \bottomrule
\end{tabular*}
\label{tab:ab_m_subset}
\end{table}

\begin{table}[ht!]
\centering
\caption{Ablation study investigating the influence of various capacity limits $K$ of the embedding dictionary.}
\begin{tabular*}{\linewidth}{@{\extracolsep{\fill}}c|cc@{\extracolsep{\fill}}}
\toprule
Capacity Limit $K$ & NDS $(\uparrow)$ & $\operatorname{mAP}$ $(\uparrow)$\\
\midrule
10 & 0.4834  & \textbf{0.3854}\\
20 & \textbf{0.4836} & 0.3839\\
30 & 0.4802 & 0.3814\\
\bottomrule
\end{tabular*}
\label{tab:abk}
\end{table}

\subsection{Visualization}

\begin{figure*}[t!]
    \centering
    \includegraphics[height=7cm,width=13cm]{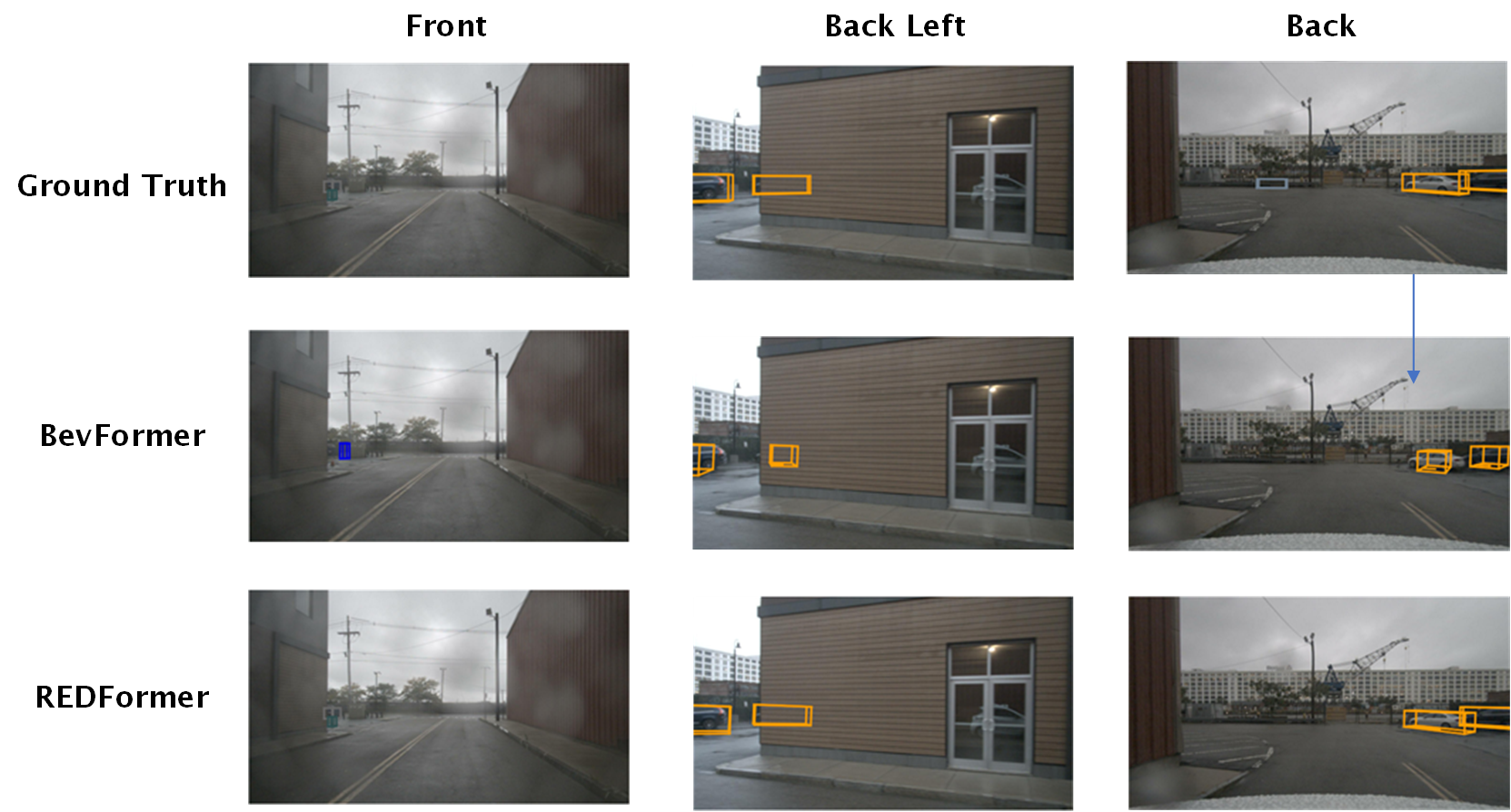}
    \caption{Visualization results of REDFormer and BEVformer on nuScenes rainy validation subset, exclusively including figures when objects are present. \textcolor{orange}{Vehicles} are marked by orange bounding boxes, \textcolor{red}{motorcycles} by red, \textcolor{blue}{pedestrians} by blue, and \textcolor{darkgray}{barriers} by gray.}
    \label{fig:visual_rain}
\end{figure*}

\begin{figure*}[t!]
    \centering
    \includegraphics[height=7cm,width=\textwidth]{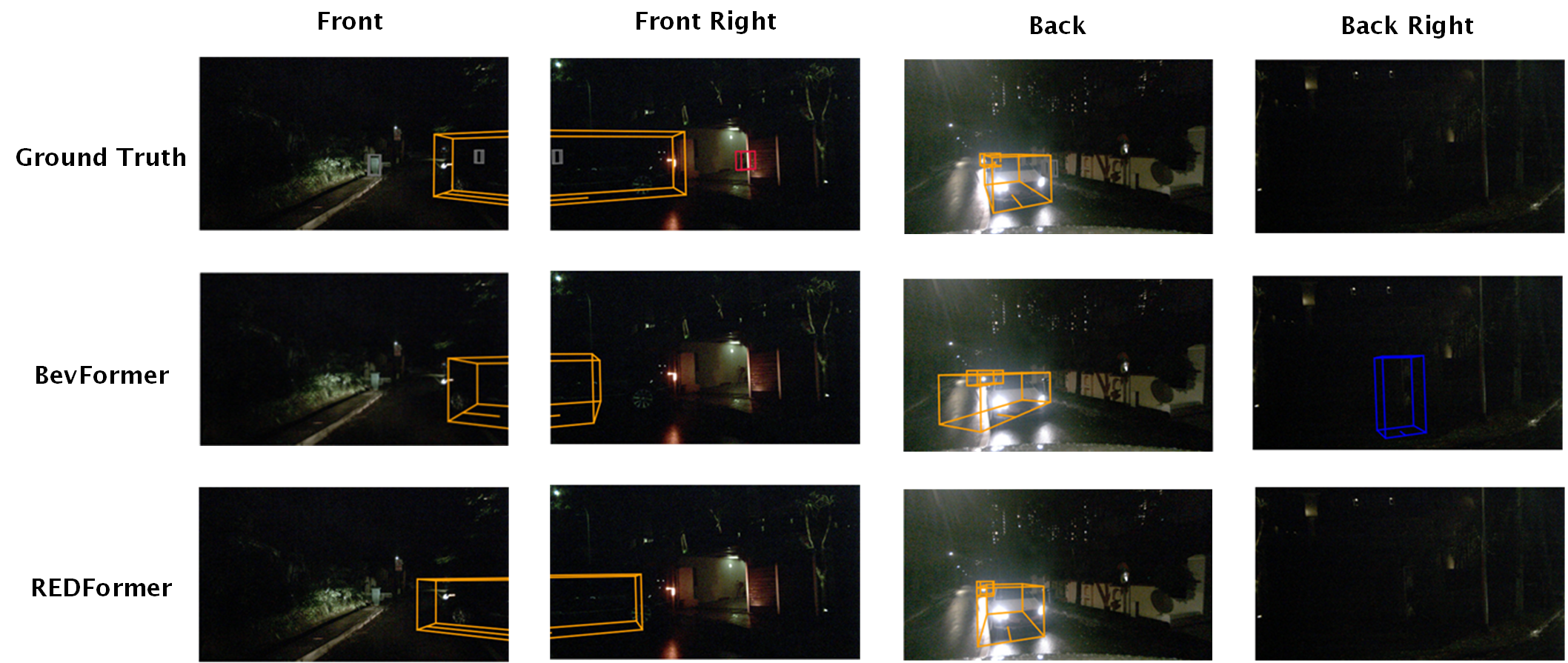}
    \caption{Visualization results of REDFormer and BEVFormer on nuScenes night validation subset, exclusively including figures when objects are present. \textcolor{orange}{Vehicles} are marked by orange bounding boxes, \textcolor{red}{motorcycles} by red, \textcolor{blue}{pedestrians} by blue, and \textcolor{darkgray}{barriers} by gray.}
    \label{fig:visual_night}
\end{figure*}

In order to provide a comprehensive insight into the performance of the REDFormer model in 3D object detection tasks under low-visibility conditions, we present detailed visualizations showcasing the outstanding performance and the significant improvement of our model compared to the baseline (BEVFormer) model in such scenarios. The comparison of prediction results between RedFormer, the baseline (BEVFormer), and ground truth in the rainy subset and night subset are visualized in Fig.\ref{fig:visual_rain} and Fig.\ref{fig:visual_night}, respectively. 

The visualization results clearly demonstrate that our model, REDFormer, outperforms the baseline model under challenging visibility conditions, as evidenced by the more precise 3D bounding boxes it generates. Notably, our model avoids entirely non-existent predictions, whereas the baseline model occasionally produces erroneous predictions. These results validate the effectiveness of our MTL approach, which enables our model to account for visibility conditions and achieve superior performance in such scenes.

Furthermore, the incorporation of multi-radar input proves to be advantageous, as it remains unaffected by environmental conditions and provides reliable cues in adverse environments. Additionally, the inclusion of radar points imparts additional depth information, resulting in more accurate and realistic 3D bounding boxes.

\section{CONCLUSIONS}

In this paper, our proposed approach to 3D object detection represents a significant improvement over existing state-of-the-art (SOTA) methods, leveraging a middle fusion technique that combines a transformer-based bird's-eye-view (BEV) encoder. We introduced an innovative radar backbone (RB) to extract features from multi-radar points and employ multi-task learning (MTL)
to enable the model to consider the impact of weather and time-of-day (TOD) on object detection. The transformer-based BEV approach enables us to effectively utilize comprehensive environmental information, leading to high performance in object detection. Our approach enhances the accuracy and robustness of the system in diverse environments, including those with reduced visibility due to adverse weather conditions or low light. By combining the benefits of MTL, the novel RB, and the transformer-based middle fusion approach, our method demonstrates significant improvements in performance. Our experiments reveal that our model outperforms the SOTA baseline model (BEVFormer), achieving a 31.31\% higher (NDS) in rainy scenes and a 46.99\% higher (NDS) in low-visibility night scenes. Overall, our approach represents a valuable contribution to the field of 3D object detection, with potential applications in a wide range of industries and use cases. One limitation of our work is that the current frames per second (FPS) of our model are relatively high. As a result, our future research will focus on optimizing the FPS and enhancing the model's deployability for real-time predictions in real automated vehicles.



\bibliography{can_1}
\bibliographystyle{IEEEtran}

\end{document}